\documentclass[10pt,twocolumn,letterpaper]{article}

\usepackage[pagenumbers]{cvpr} 

\definecolor{cvprblue}{rgb}{0.21,0.49,0.74}
\usepackage[pagebackref,breaklinks,colorlinks,allcolors=cvprblue]{hyperref}

\usepackage{xcolor,amssymb}
\newcommand{\cmark}{\textcolor{green!60!black}{\checkmark}}
\newcommand{\xmark}{\textcolor{red}{\text{\sffamily X}}}

\usepackage[accsupp]{axessibility}  
\usepackage[T1]{fontenc}

\def\paperID{} 
\def\confName{CVPR}
\def\confYear{2026}

\title{A Polarized Reflection and Material Dataset of Real World Objects}

\author{Jing Yang, Krithika Dharanikota, Emily Jia, Haiwei Chen, Yajie Zhao\\
Institute for Creative Technologies,
University of Southern California\\
{\tt\small \{jyang, kdharanikota, ejia, chenh, zhao\}@ict.usc.edu}
}

\begin{document}
\maketitle

\begin{figure*}[ht]
    \centering
    \vspace{-10pt}
    \includegraphics[width=1\linewidth]{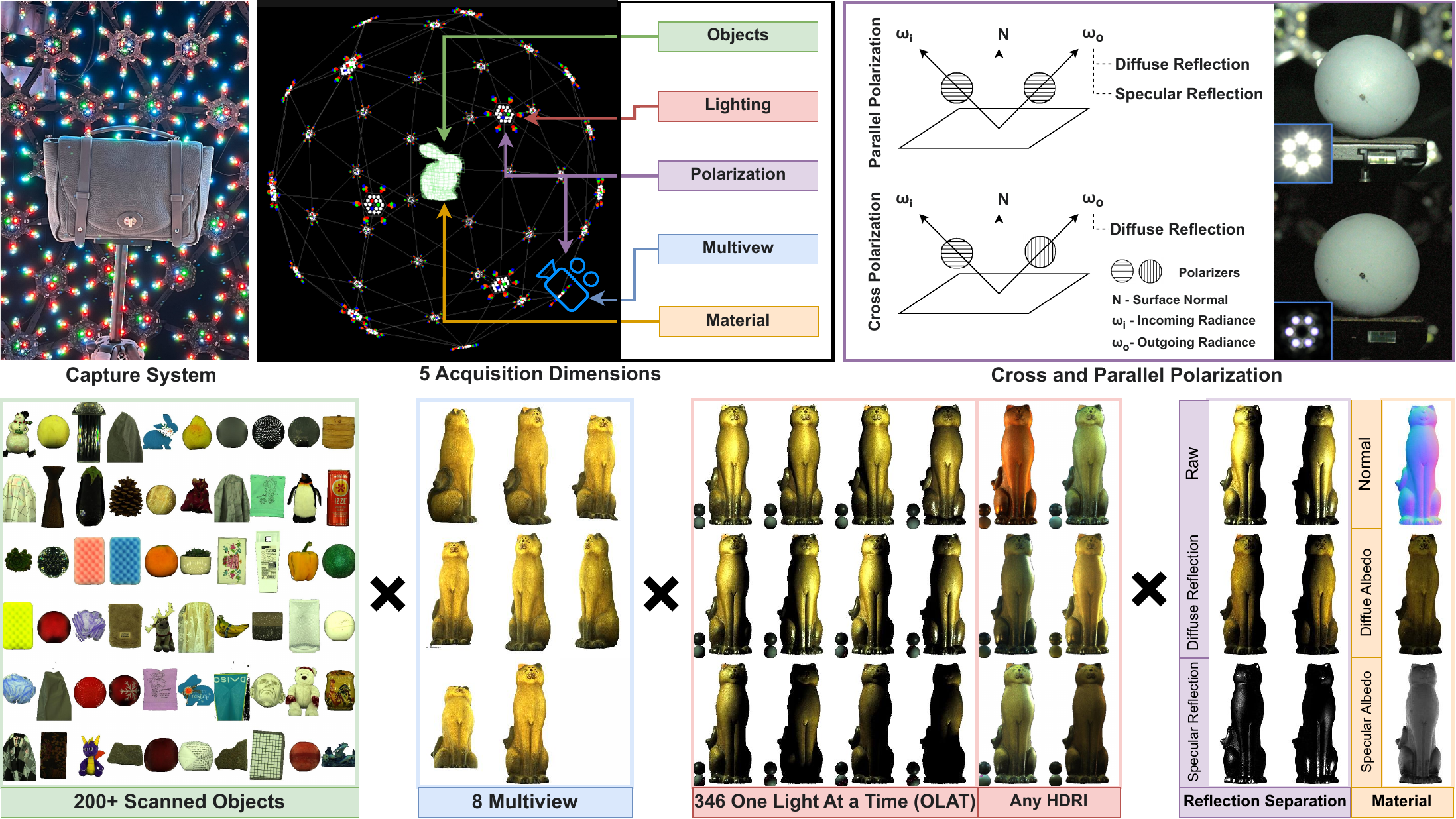}
    \caption{\textbf{Overview of our polarized reflection and material dataset.} Our Light Stage capture system records real-world objects under controlled lighting, polarization, and multiview conditions. The dataset spans five acquisition dimensions: objects, lighting, polarization, multiview, and material. Each object is captured under 346 One-Light-at-a-Time (OLAT) illuminations and can be synthesized into arbitrary HDRI lighting. Cross and parallel polarization enable reflection separation into diffuse and specular components, from which material attributes such as albedo and normal are derived, resulting in more than 1,200,000 (1.2 million) captured images at 6144×3240 resolution.}
    \label{fig:capture_system_and_dataset}
\end{figure*}

\begin{abstract}

Accurately modeling how real-world materials reflect light remains a core challenge in inverse rendering, largely due to the scarcity of real measured reflectance data. Existing approaches rely heavily on synthetic datasets with simplified illumination and limited material realism, preventing models from generalizing to real-world images. We introduce a large-scale polarized reflection and material dataset of real-world objects, captured with an 8-camera, 346-light Light Stage equipped with cross/parallel polarization. Our dataset spans 218 everyday objects across five acquisition dimensions—multiview, multi-illumination, polarization, reflectance separation, and material attributes—yielding over 1.2M high-resolution images with diffuse–specular separation and analytically derived diffuse albedo, specular albedo, and surface normals. Using this dataset, we train and evaluate state-of-the-art inverse and forward rendering models on intrinsic decomposition, relighting, and sparse-view 3D reconstruction, demonstrating significant improvements in material separation, illumination fidelity, and geometric consistency. We hope that our work can establish a new foundation for physically grounded material understanding and enable real-world generalization beyond synthetic training regimes. Project page: \href{https://jingyangcarl.github.io/ICTPolarReal/}{jingyangcarl.github.io/ICTPolarReal}
\end{abstract}    
\section{Introduction}








Modeling how physical materials reflect light is a cornerstone that transforms our understanding of natural images. Typically, the image capturing process that intricately entangles geometry, light sources and material properties is considered non-invertible. However, past decades have seen extensive efforts from computer vision research to estimate intrinsics images from natural photos, a process known as intrinsic image decomposition~\cite{barrow1978recovering} or inverse rendering~\cite{ramamoorthi2001signal}. While traditional methods towards this end had to rely on hand-crafted priors for estimation of low-order effects~\cite{chen2013simple, garces2012intrinsic, zhao2012closed, rother2011recovering}, the quality of intrinsic image decomposition has been drastically improved in recent years thanks to the emergence of dedicated synthetic datasets~\cite{objaverse, roberts2021hypersim, li2018cgintrinsics, li2021openrooms} and large generative models~\cite{zeng2024rgbx,luo2024intrinsicdiffusion,liang2025diffusion,chen2025uni,choi2025channel,zheng2025dnf,sun2025ouroboros}. 

The majority of improvements observed in recent intrinsic image decomposition models stem from synthetically rendered data. While CGI renderers can produce visually appealing images, they are constrained by simplified illumination models and limited material realism. Common shading models approximate bidirectional reflectance using analytical BRDFs or a small number of samples, ignoring multi-scattering, polarization, and subsurface transport that are ubiquitous in real objects. Moreover, few synthetic assets possess physically validated material parameters or polarization information, leading to discrepancies between rendered and real observations. Consequently, models trained exclusively on synthetic data often fail to generalize to uncontrolled lighting or real photographs. Recently, \cite{murmann2019dataset} collected the Multi-Illumination Dataset that provides real-life photos captured in different lighting conditions, but their lack of measurement for intrinsic decompositions limit their uses in training supervised image decomposition generative models.

In this paper, we tackle the challenges to collect a real-life reflection and material dataset, where we physically measured the surface reflectance of apples, bananas, cantaloupes, cabbages, onions, sweet potatoes, fabric bags, handbag logos, reusable bags, metal kettles, marbles, mirror balls, ceramic jars, glass bottles, plastic toys, rubber balls, wooden figurines, paper boxes, stone pots, and many more, covering a broad spectrum of material types including organic produce, plastics, rubbers, metals, ceramics, glass, fabrics, wood, paper, and other everyday household materials. By exploiting polarization, our system physically separates diffuse and specular reflections under controlled One-Light-at-a-Time (OLAT) illumination. Each object is observed from eight viewpoints and 346 lighting directions, yielding dense, polarization-separated reflectance fields. The dataset covers a diverse set of everyday objects, ranging from metals and ceramics to translucent and rough dielectrics, each captured with high dynamic range and geometric calibration. This results in the first real-life dataset that can directly supervise deep neural networks that perform material decomposition and physically-based neural rendering.

To validate the effectiveness of our constructed dataset, we conduct comprehensive qualitative and quantitative evaluations of state-of-the-art techniques across multiple tasks, including intrinsic image decomposition, forward relighting, and sparse-view 3D reconstruction. These evaluations demonstrate that models trained or fine-tuned on our real-world measurements achieve markedly improved material separation, more faithful lighting response, and greater robustness to real-world appearance variations, capabilities that synthetic-only training cannot provide.

\section{Related Works}

\subsection{Inverse Rendering} 

Originally proposed by ~\cite{barrow1978recovering}, the inverse rendering problem, also known as intrinsic image decomposition, aims to separate an image into albedo, irradiance, and optionally, a specular component. Traditional methods address this problem by imposing prior assumptions on the different components, involving smooth shading ~\cite{chen2013simple, garces2012intrinsic}, grayscale shading~\cite{zhao2012closed,garces2012intrinsic}, albedo sparsity~\cite{garces2012intrinsic,bell2014intrinsic,rother2011recovering,meka2021real}, and others~\cite{choi2023ibl,careaga2023intrinsic}.  Recently, with the rapid development of deep learning, many studies have sought to learn such priors directly from data. Both supervised approaches~\cite{zhou2015learning,fan2018revisiting,shi2017learning,narihira2015direct,boss2020two} based on RGB renderings with ground-truth decompositions or annotations, and unsupervised methods~\cite{wang2021learning,li2018cgintrinsics} using only multi-illumination images have been explored. 

Many recent works have posed inverse rendering as an image-to-image translation problem that leverages rich priors from large image generation models~\cite{zeng2024rgbx,luo2024intrinsicdiffusion,liang2025diffusion,chen2025uni,choi2025channel,zheng2025dnf,sun2025ouroboros}. While \cite{bhattad2023stylegan} has shown that an image generation model trained without any intrinsic image still can obtain implicit understanding of intrinsic image decomposition, many other works have relied on synthetic dataset that provides mapping from images to their decomposed components and mainly varied in the use of illumination representations (e.g. irradiance image in~\cite{zeng2024rgbx}, HDR environment map in~\cite{liang2025diffusion}, Phong illumination model in~\cite{wimbauer2022rendering}) and the selection and processing of synthetic datasets. In addition, large foundation models~\cite{he2025lotus, ke2025marigold} have specifically targeted the estimation of image depth and normal.


\subsection{Dataset for Inverse Rendering} 
Due to the difficulty of measuring reflectance in real life, the majority of existing datasets are synthetically modeled and rendered by computer graphics programs. Objaverse~\cite{objaverse, objaverseXL} provides a large collection of 3D models, part of which contains materials that can be used to create intrinsics images. Hyperism~\cite{roberts2021hypersim},  CGIntrinsic~\cite{li2018cgintrinsics}, OpenRooms~\cite{li2021openrooms}, InvIndoor~\cite{li2020inverse} and InteriorNet~\cite{li2018interiornet} are synthetic indoor scenes that provide ground truth material decomposition. Multi Illumination~\cite{murmann2019dataset} provides real-life indoor photos under various illumination conditions with material annotations. However, the dataset is not suitable for direct supervision of inverse rendering models due to its lack of measurements for intrinsic images corresponding to the captured photos. 

OpenSVBRDF~\cite{10.1145/3618358} acquires 1000 spatially varying reflectance flat samples with a acquiistion device, but the data is restricted to near-planar samples and only two viewpoints, limiting its applicability to full 3D objects and multi-view reconstruction. Open Illumination~\cite{liu2024openillumination} instead focuses on illumination variation and captures only a small number of objects under a limited set of lighting patterns, providing valuable but narrow coverage of object and material diversity.

\section{Dataset}

\begin{table*}
    \centering
    \scriptsize  
    \setlength{\tabcolsep}{2pt} 
    \begin{tabular}{c|llcclll|cl|llll}
 &    \multicolumn{7}{c|}{Capture}&\multicolumn{2}{c|}{Lighting} & \multicolumn{4}{c}{Reflection Reflection}\\
         Dataset&   Type& Real-World &Geometry&  Albedo&      Specular&Normal&Multiview&HDRI&OLAT & Polarization&diffuse only&specular only&mix\\
         \hline
 Objaverse~\cite{objaverse}&  Objects& \xmark&\cmark& \cmark&      \cmark&\cmark&\cmark&\cmark& \cmark&  \xmark&\cmark& \cmark&\cmark\\
 Hypersim~\cite{roberts2021hypersim}&  Indoor Scene& \xmark&\cmark& \cmark&   \cmark&\cmark&\cmark& \cmark& \xmark&  \xmark&\xmark& \xmark&\cmark\\
 Interverse~\cite{zhu2022learning}&  Indoor Scene& \xmark&\cmark& \cmark&      \cmark&\cmark&\cmark&\cmark& \xmark&  \xmark&\xmark& \xmark&\cmark\\
 Multi Illumination\cite{murmann2019dataset}&  Objects& \cmark&\xmark& \xmark&      \xmark&\xmark&\xmark&\xmark& \xmark&  \xmark&\xmark& \xmark&\cmark\\
D. Luo et al \cite{Luo:2024:Correlation-aware}&  Flat Surface& \cmark&\xmark& \cmark&   \cmark&\cmark&\xmark& \xmark& \cmark& \xmark& \xmark& \xmark&\cmark\\
 OpenSVBRDF\cite{10.1145/3618358}& Flat Surface& \cmark& \xmark& \cmark& \cmark& \cmark& \xmark& \xmark& \cmark& \xmark& \xmark& \xmark&\cmark\\
         Ours&   Objects& \cmark&\cmark&  \cmark&       \cmark&\cmark&\cmark&\cmark& \cmark&  \cmark&\cmark& \cmark&\cmark\\
    \end{tabular}
    \caption{\textbf{Comparison between Material and Lighting-related Datasets.} Ours is the first real-world dataset that provides both comprehensive lighting coverage, polarization and material attributes enabling direct supervision for Inverse Rendering and Relighting.}
    \label{tab:dataset_feature}
\end{table*}


We target the acquisition of \textbf{real-world, everyday objects} spanning a wide range of materials such as metals, ceramics, plastics, fabrics, wood, leather, glass, and translucent organics. In order to disentangle diffuse and specular reflections, each object is captured as a multi-view, polarized reflectance field that records how light interacts with the surface under hundreds of controlled illumination directions. From these measurements, we derive physically interpretable quantities including diffuse and specular reflection components, surface normal, and material albedo. This dataset provides the foundation for learning and evaluating physically grounded appearance models under real-world conditions. Examples are shown in Figure~\ref{fig:capture_system_and_dataset}.



\subsection{Capture System} The capture system consists of 346 LED light sources mounted on a geodesic dome and 8 synchronized RED Komodo 6K global shutter cameras positioned to cover a broad angular range. Each LED is equipped with either a cross- or parallel-oriented linear polarizer arranged in a hexagonal tiling pattern. The cameras are also equipped with rotatable linear polarizers, enabling consistent acquisition of polarized reflectance. For each lighting direction, we acquire two polarized observations per view: one under cross-polarized lighting and one under parallel-polarized lighting. This configuration enables per-light, per-view separation of diffuse and specular reflections via linear polarization analysis. Lights are triggered in a spiral order, \textbf{One Light At A Time (OLAT)}, from the frontal to the back hemisphere to ensure angularly uniform coverage. The full sequence consists of 346 lighting conditions captured across 8 viewpoints at 6144×3240 resolution for each polarization status.

\subsection{Acquisition} 
Our data acquisition consists of two stages: (1) polarized image capture and (2) material processing.
The first stage records polarization-separated reflectance under controlled illumination, and the second computes physically based material parameters.

\paragraph{Polarized Image Capture} 
To capture polarized image, we first perform system calibration that covers lighting, camera, and color calibration to ensure accurate lighting directions, view geometry, and consistent color response across all cameras for the subsequent optimization. The change in lighting polarization states is achieved by selectively activating different light sources on the light board. Each board adopts a hexagonal LED layout with alternating cross- and parallel-oriented polarizers in front of adjacent lights. This design enables capture of two complementary polarized images per illumination direction: cross-polarized ($I_\perp$) and parallel-polarized ($I_\parallel$) as shown in Fig.~\ref{fig:capture_system_and_dataset}.

In theory, placing linear polarizers in front of light sources and analyzers with varying orientations before the camera enables the separation of diffuse and specular reflections. The intensity follows Malus’s Law: $I = I_0 \cos^2\theta$,
where $\theta$ denotes the angle between the transmission axes of the polarizer and analyzer. Since the mean value of $\cos^2\theta$ over all orientations is $\tfrac{1}{2}$, the diffuse and specular radiance components under identical illumination can be expressed as:
\begin{equation}
I_d = 2I_\perp, \quad I_s = 2I_\parallel - 2I_\perp,
\end{equation}

During capture, each object is scanned under cross- and parallel-polarized OLAT illumination, producing image sequences for all 346 lighting directions in 8 views:
\begin{equation}
    \Lambda_\perp^{i}=\{I_\perp^k\}_{k=0}^{N}, \Lambda_\parallel^{i}=\{I_\parallel^k\}_{k=0}^{N}, N=346.
\end{equation}

\paragraph{Material Processing} 
We present each material using diffuse albedo $\rho_d$, specular albedo $\rho_s$, and surface normal $n$. From the polarized images, diffuse and specular reflection sequences are computed as: 
\begin{equation}
    \Lambda_d=2\Lambda_\perp, \quad \Lambda_s=2(\Lambda_\parallel-\Lambda_\perp).
\end{equation}

Following Lambert’s cosine law, we obtain the diffusion reflection by
\begin{equation}
\label{eqn:lambert-cosine-law}
\begin{split}
L_o = \frac{L_i(\omega_i\cdot n)(\omega_o\cdot n) d\Omega dA}{d \Omega_o (\omega_o \cdot n) d A_o}=\frac{L_i(\omega_i \cdot n) d\Omega dA}{d\Omega_o dA_o}
\end{split}
\end{equation}
where $L_i$ is the incident radiance from the light source and $L_o$ denotes the observed radiance. For each pixel of aperture area \(dA_o\), diffuse albedo $\rho_d$ and surface normal $n$ are solved by minimizing $L=\{\rho_d | n\cdot\omega_k|\}^N_{k=0}-\Lambda_d$.

We then derive the specular albedo $\rho_s$ via the following:
\begin{equation}
\rho_s=\int_\Omega R_s(\omega_i, \omega_o) d\omega_i \approx \frac{4\pi\kappa}{N} \sum\nolimits_{k=0}^N \hat{I}_s^k,
\end{equation}
where $R_s$ represents the specular reflection function of the material. $\kappa$ is a constant determined by light intensity and solid angle in the capture device.

This two-stage process yields polarization-separated, multiview reflectance fields with corresponding material parameters suitable for physically-based rendering and relighting.

\subsection{Dataset Statistics} Our dataset comprises 218 objects scanned across 346 lighting directions, 2 lighting polarization status and 8 views, totaling over 1.2M high-dynamic-range images at 6K resolution. For each object, we provide the raw cross- and parallel-polarized image sequences, along with the computed reflectance maps. The scanned objects span a diverse range of real-world materials, including metals, plastics, ceramics, concrete, fabrics, leathers, organics, wood, and transparent or translucent materials such as glass and wax. We further highlight the differences of our data with the existing dataset in Table~\ref{tab:dataset_feature}.

\paragraph{Lighting Augmentation} \label{sec:lighting-augmentation} Extending from the OLAT capture setup, our dataset supports the synthesis of arbitrary novel illuminations with corresponding diffuse and specular separation. Given any environmental lighting texture as an environment map, we project it onto a unit sphere aligned with the calibrated light directions. For each light source, we sample the surrounding pixels on the environment map and compute their average intensity to determine the weighting of that light. The final relit result is obtained by performing a weighted sum over all OLAT images using these computed weights. This approach enables infinite lighting augmentation while maintaining accurate diffuse and specular ground truth through polarized imaging.

\begin{figure*}[h]
    \centering
    \vspace{-10pt}
    \includegraphics[width=1\linewidth]{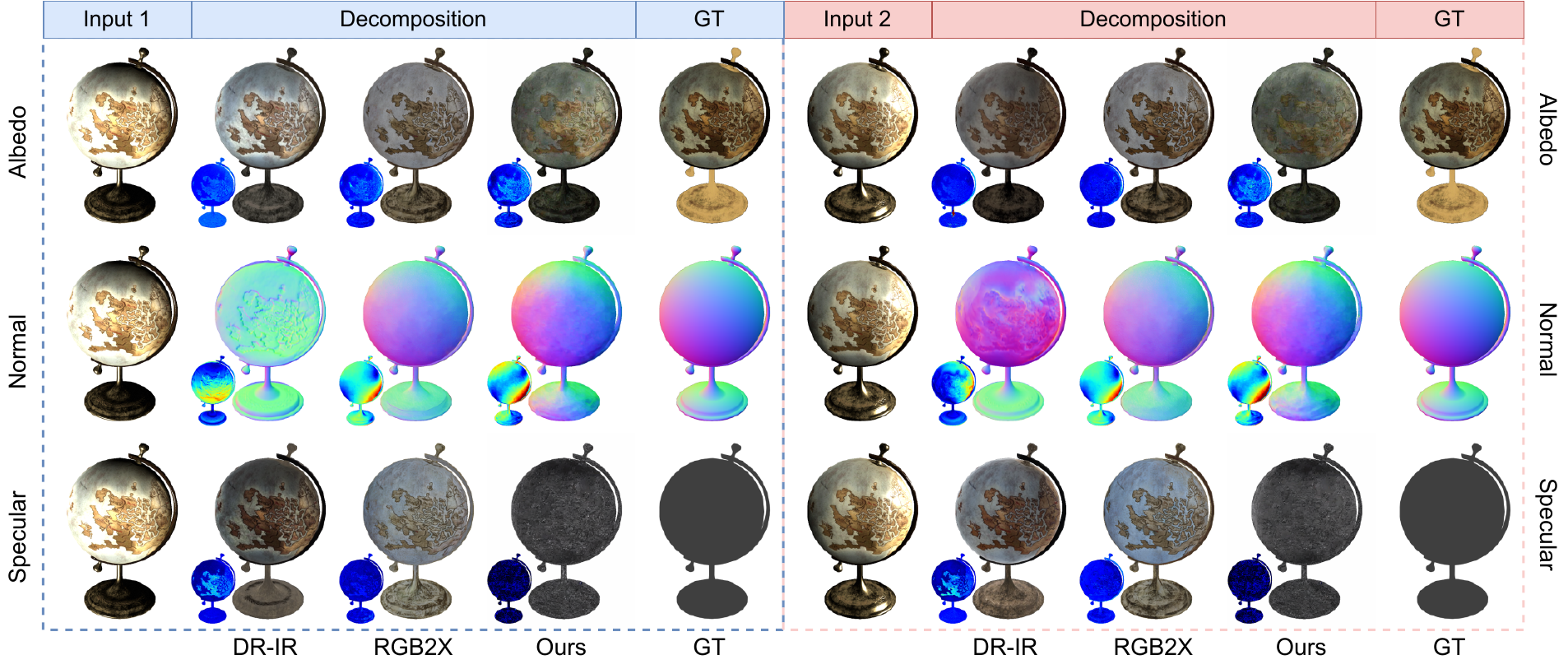}
    \caption{\textbf{Decomposition results on Objaverse samples}. Each row shows the predicted albedo, normal, and specular components from DR-IR, RGB2X, and ours under two lighting conditions. An error map is shown in the lower-left corner of each result. Post-training with our dataset produces more consistent decompositions and smoother material separation across illumination changes.}
    \label{fig:inverse-objaverse}
\end{figure*}

\begin{table*}[t]
    \centering
    \small
    \setlength{\tabcolsep}{3pt}
    \begin{tabular}{cc|cccc|cccc|cccc}
        \toprule
        & & \multicolumn{4}{c}{Albedo} & \multicolumn{4}{c}{Normal} & \multicolumn{4}{c}{Specular} \\
        Baseline & Dataset & MSE & PSNR & SSIM & LPIPS & MSE & PSNR & SSIM & LPIPS & MSE & PSNR & SSIM & LPIPS \\
        \midrule
        \multicolumn{14}{c}{\textbf{HDRI Lighting}} \\
        \midrule
        lotus-D~\cite{he2025lotus}  & lightstage & -- & -- & -- & -- & 0.010 & 24.722 & 0.767 & 0.398 & -- & -- & -- & -- \\
        dsine~\cite{bae2024dsine}  & lightstage & -- & -- & -- & -- & 0.008 & 25.509 & \textbf{0.776} & 0.401 & -- & -- & -- & -- \\
        DR~\cite{liang2025diffusion}  & lightstage & 0.035& 20.011& 0.807& 0.404& 0.024& 20.484& 0.663& 0.623& 0.041 & 22.614 & \textbf{0.846} & 0.362 \\
        rgb2x~\cite{zeng2024rgbx}  & lightstage & 0.040 & 18.075 & 0.812 & 0.466 & 0.037 & 18.576 & 0.679 & 0.702 & 0.047 & 17.210 & 0.791 & 0.495 \\
        ours & lightstage & \textbf{0.005} & \textbf{33.509} & \textbf{0.935} & \textbf{0.198} & \textbf{0.005} & \textbf{28.086} & 0.698 & \textbf{0.400} & \textbf{0.002} & \textbf{31.016} & 0.775 & \textbf{0.338} \\
        \midrule
        DR~\cite{liang2025diffusion}  & Objaverse & 0.048 & 23.206& 0.855& \textbf{0.255}& 0.018 & 25.096 & 0.895 & 0.306 & 0.191 & 15.106 & 0.868 & 0.274 \\
        rgb2x~\cite{zeng2024rgbx} & Objaverse & 0.042& 22.882 & 0.888 & 0.333 & 0.008 & 28.825 & 0.886 & 0.358 & 0.052 & 20.278 & 0.860 & 0.451 \\
        ours & Objaverse & \textbf{0.032}& \textbf{26.691}& \textbf{0.899}& 0.289 & 0.012 & 26.971 & 0.904 & 0.305 & 0.177 & 15.181 & 0.733 & 0.497 \\
        \midrule
        \multicolumn{14}{c}{\textbf{One-Light-at-a-Time (OLAT) Lighting}} \\
        \midrule
        lotus-D~\cite{he2025lotus} & lightstage & -- & -- & -- & -- & 0.019 & 20.914 & \textbf{0.771} & 0.500 & -- & -- & -- & -- \\
        dsine~\cite{bae2024dsine} & lightstage & -- & -- & -- & -- & 0.021 & 21.220 & 0.764 & 0.489 & -- & -- & -- & -- \\
        DR~\cite{liang2025diffusion} & lightstage & 0.030 & 20.264 & \textbf{0.778} & 0.450 & 0.020 & 21.084 & 0.634 & 0.608 & 0.024 & 26.593 & 0.842 & 0.360 \\
        rgb2x~\cite{zeng2024rgbx} & lightstage & 0.050 & 16.957 & 0.757 & 0.525 & 0.022 & 20.345 & 0.690 & 0.574 & 0.059 & 16.334 & 0.782 & 0.491 \\
        ours & lightstage & \textbf{0.017} & \textbf{23.844} & 0.737 & 0.460 & \textbf{0.009} & \textbf{25.206} & 0.637 & 0.597 & \textbf{0.011} & 24.544 & 0.659 & 0.509 \\
        \midrule
        DR~\cite{liang2025diffusion} & Objaverse & 0.029 & 25.470 & 0.812 & 0.298 & 0.018 & 25.087 & 0.839 & 0.364 & 0.165 & 16.009 & 0.914 & 0.180 \\
        rgb2x~\cite{zeng2024rgbx} & Objaverse & \textbf{0.021} & 25.162 & 0.657 & 0.578 & 0.012 & 26.971 & \textbf{0.904} & \textbf{0.305} & 0.081 & 18.247 & 0.715 & 0.623 \\
        ours & Objaverse & 0.040 & \textbf{25.914} & \textbf{0.929} & \textbf{0.264} & \textbf{0.010} & \textbf{27.685} & 0.600 & 0.347 & 0.169 & 15.361 & 0.696 & 0.507 \\
        \bottomrule
    \end{tabular}
    \caption{Benchmarking decomposition under HDRI and OLAT lighting. Each lighting type is shown in a separate block for clarity. 
    }
    \label{tab:inverse_evaluation_gbuffer}
\end{table*}

\begin{figure}[h]
    \centering
    \vspace{-10pt}
    \includegraphics[width=1\linewidth]{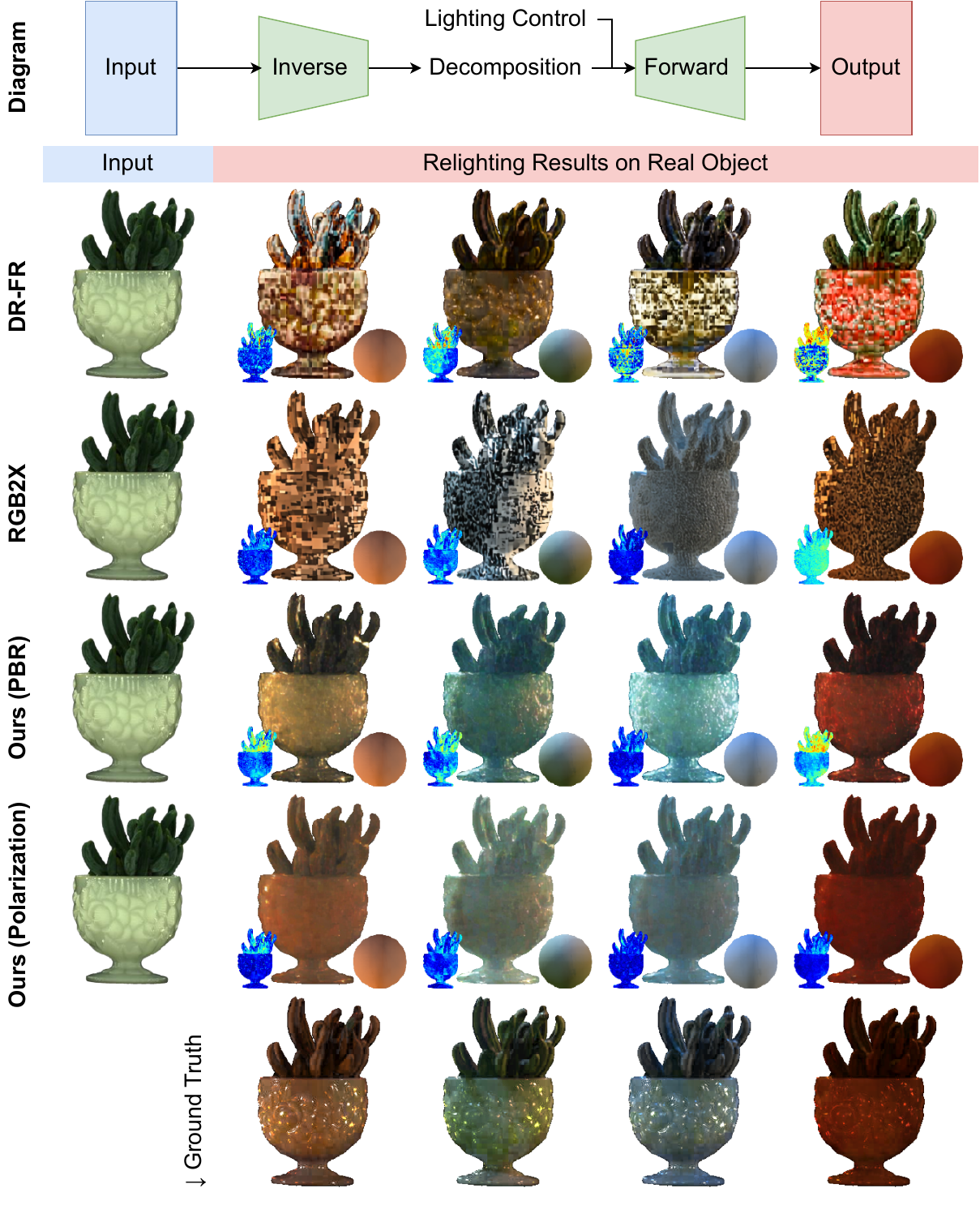}
    \caption{\textbf{Forward relighting results on real objects (Light Stage dataset) under HDRI lighting} Each row shows the relighting outputs from DR-FR, RGB2X, and our model under both the PBR and polarization workflows. The top diagram illustrates the full inverse-to-forward relighting pipeline. For each relit result, the error map is shown in the lower-left corner and the lighting reference in the lower-right corner.}
    \label{fig:relighting-lightstage}
\end{figure}

\begin{figure}[h]
    \centering
    \vspace{-10pt}
    \includegraphics[width=1\linewidth]{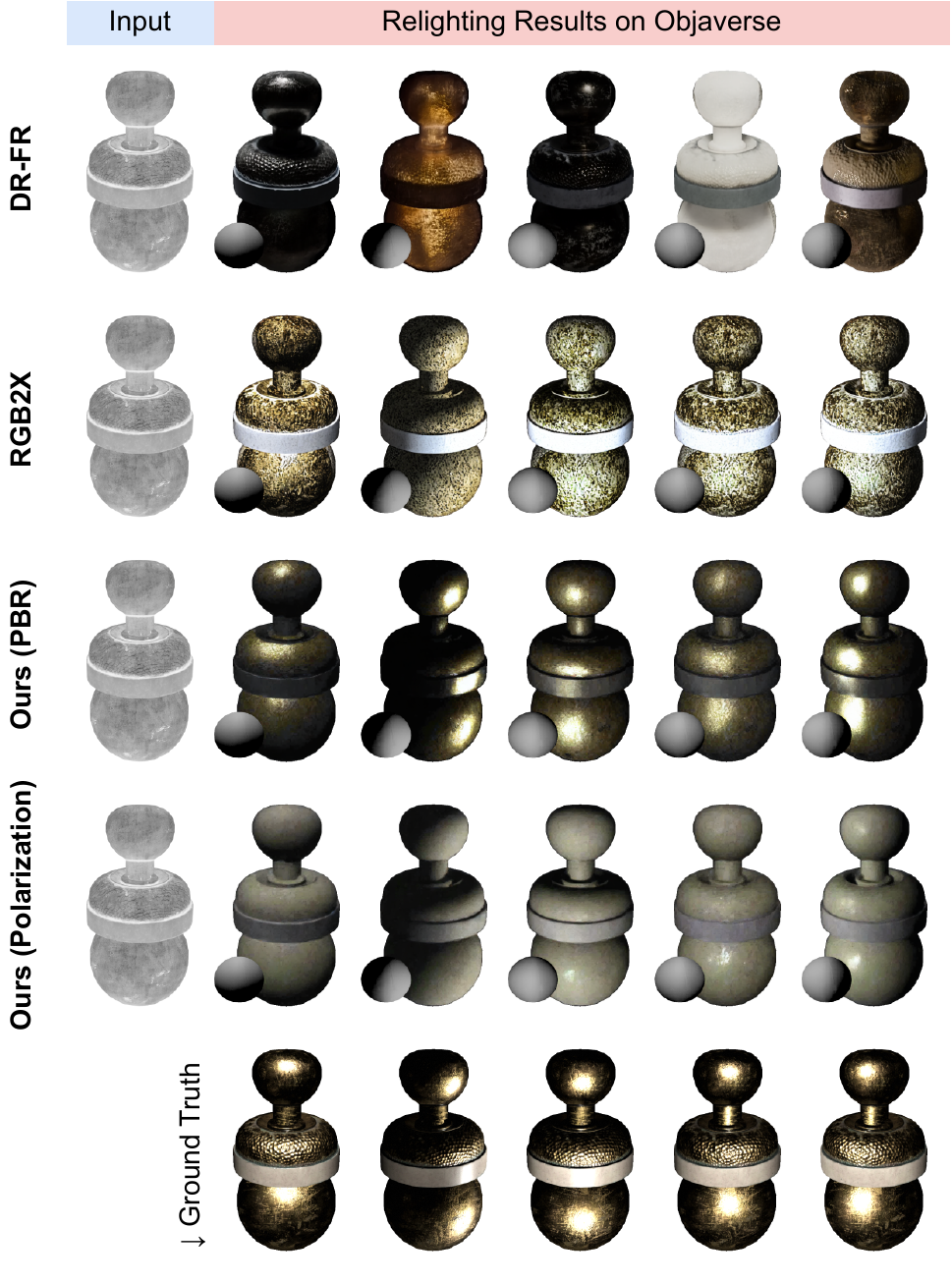}
    \caption{\textbf{Forward relighting results on the Objaverse dataset under OLAT lighting.} The lighting reference is shown in the lower-left corner. }
    \label{fig:relighting-objaverse}
\end{figure}

\begin{figure}[h]
    \centering
    \vspace{-10pt}
    \includegraphics[width=1\linewidth]{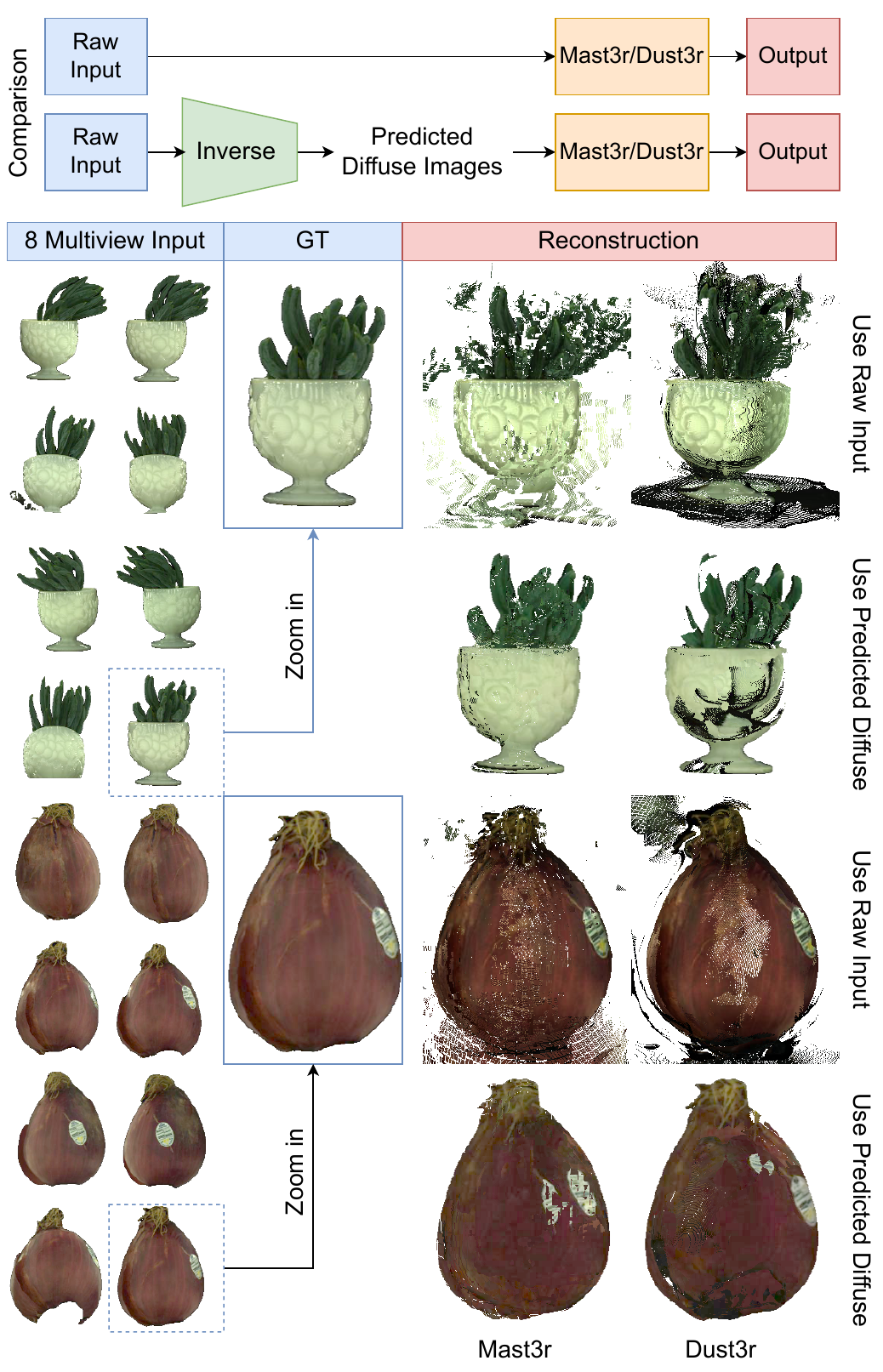}
    \caption{\textbf{Impact of our inverse model on sparse-view 3D reconstruction.} We compare reconstructions from Dust3r and Mast3r using raw input images and our predicted diffuse images from the inverse decomposition model. The reconstruction takes 8 sparse multiview input.}
    \label{fig:reconstruction-lightstage}
\end{figure} 

\section{Inverse and Forward Rendering Model}

To evaluate the effectiveness of the constructed dataset, we utilize our dataset to train both inverse and forward neural rendering models. Specifically, we treat both neural rendering as an image-to-image translation problem and finetune from RGB2X~\cite{zeng2024rgbx}, an image decomposition and synthesis method pretrained on synthetic data. For both training, we employ Low-Rank Adaptation~\cite{hu2022lora} to prevent catastrophic forgetting. 

\paragraph{Inverse Rendering Network.} 
We design two types of inverse rendering models that produce decomposed intrinsic information in Physically-based Rendering (PBR) representation and Polarization representation. 

\textbf{Physically-based Rendering Workflow.} Following the graphics convention, we design our decomposition network to predict physically meaningful material components, including diffuse albedo, specular albedo, and surface normal. The architecture is compatible with recent diffusion-based relighting frameworks, enabling seamless integration with existing pipelines. 

\textbf{Polarization Workflow.} In addition to the traditional physically-based decomposition, we introduce a polarization-based workflow. The model decomposes the input into polarization components corresponding to cross-polarized and parallel-polarized images.
Although diffuse and specular separation through polarization is not commonly available in standard imaging setups, polarization provides a powerful physical cue for reflection disentanglement. Specular reflections often disrupt multiview correspondences in 3D reconstruction due to their view-dependent nature, leading to geometry inconsistencies and reconstruction artifacts. By training a model to “virtualize” polarization, predicting cross- and parallel-polarized components from unpolarized inputs, we can recover diffuse and specular layers that facilitate more accurate relighting and benefit other challenging vision tasks that rely on precise reflectance modeling.

To control the diffusion model output in both workflows, we adopt a prompt-based conditioning mechanism: each target component (for example, “albedo” or “surface normal”) is encoded into a text embedding and injected into the diffusion model to guide the generation process. 

\paragraph{Forward Rendering Network.} We additionally train a forward rendering network with our dataset to render images from the intrinsic components, represented in either physically-based rendering (PBR) or polarization format. To explicitly model light interaction, we introduce an additional irradiance map computed as the integrated $\sum (n \cdot \mathbf{\omega})$ over all sampled light directions weighted by the corresponding light source color. 

During training, we employ lighting augmentation that combines three types of input images: (1) real captures under one-light-at-a-time (OLAT) illumination, (2) synthesized HDRI lighting generated via weighted summation of OLAT images using environment-map-based light weights (Sec.~\ref{sec:lighting-augmentation}), and (3) direct captures under all-white illumination. For each training sample, For each training sample, we randomly select an object’s intrinsic material components, including diffuse albedo, surface normal in camera space, and specular albedo from an arbitrary camera view, and pair them with the associated irradiance map as inputs. The text prompt is left empty, and supervision is applied through an L2 loss with respect to the target image under the corresponding augmented lighting condition.


\section{Evaluation and Experiments}

\begin{table*}
    \centering
    \small  
    \setlength{\tabcolsep}{2pt} 
    \vspace{-10pt}
    \begin{tabular}{@{}cc|cccl|cccl}
 & Input Lighting& \multicolumn{4}{c|}{HDRI}& \multicolumn{4}{c}{OLAT}\\
 \hline
         Baseline&   dataset& MSE& PSNR& SSIM &LPIPS& MSE& PSNR& SSIM &LPIPS \\
         \hline
         
         DR\cite{liang2025diffusion} &   lightstage&0.058& 16.972& 0.775&0.386& 0.020& 23.565&  0.798&0.335 \\
         
         rgb2x\cite{zeng2024rgbx} &   lightstage&0.038& 18.495& 0.514&0.514& 0.046& 17.529&  0.475&0.559 \\
         
  ours-GBuffer & lightstage& \textbf{0.005}& \textbf{27.798} & \textbf{0.904}& 0.211 & 0.011 & 24.850 & \textbf{0.850}&\textbf{0.301} \\
   ours-Polarization & lightstage& 0.007& 26.133 & 0.909 & \textbf{0.200}& \textbf{0.010}& \textbf{25.331}& 0.802 & 0.388\\
 \hline
 DR\cite{liang2025diffusion} & Objaverse& 0.066& 20.934& 0.754& 0.517 & 0.026& 25.092& 0.873& 0.229 \\
         rgb2x\cite{zeng2024rgbx} &   Objaverse&0.032& 24.386&  0.595&0.606 & 0.014& 26.861&   0.598&0.547 \\
 ours-GBuffer & Objaverse& 0.014& 26.762 & \textbf{0.956}& 0.103 & \textbf{0.014}& \textbf{27.638}& \textbf{0.932}& \textbf{0.203}\\
  ours-Polarization & Objaverse& \textbf{0.007}& \textbf{29.554}& 0.964 & \textbf{0.093}& 0.021 & 25.689 & 0.914 & 0.249\\
    \end{tabular}
    \caption{Evaluation on relighting.}
    \label{tab:forward_evaluation}
\end{table*}

\begin{table*}
    \centering
    \small
    \setlength{\tabcolsep}{2pt} 
    \begin{tabular}{|c|cccl|cccl|llll|}
    \hline
 Input Settings &\multicolumn{4}{c|}{Raw Images
}& \multicolumn{4}{c|}{Pred Diffuse Images
}& \multicolumn{4}{c|}{Pred Albedo Images}\\
\hline
         Methods&MSE&  PSNR&  SSIM  &LPIPS
&  MSE&  PSNR& SSIM &LPIPS
& MSE& PSNR&SSIM  &LPIPS\\
         \hline
         Dust3r &0.102&  14.512&  0.226 &0.604
&  0.060&  17.783&  0.411 &0.556
& \textbf{0.054}& \textbf{20.302}&\textbf{0.513} &\textbf{0.506}\\

\hline
         Mast3r &0.161&  12.719&  0.193 &0.613
&  0.132&  13.777&  0.247 &0.615
& \textbf{0.130}& \textbf{15.565}&\textbf{0.282} & \textbf{0.603}\\
\hline
    \end{tabular}
    \caption{\textbf{Sparse-view 3D reconstruction on 50 real-world objects (8-view input).}  We compare four input settings: raw, polarized diffuse, predicted diffuse (polarization workflow), and predicted albedo (PBR workflow). 
Both methods are evaluated with and without masking. The best score in each row across settings is highlighted in bold. 
    }
    \label{tab:sparse-view-reconstruction}
\end{table*}

\subsection{Setup}

\paragraph{Baselines} Lotus~\cite{he2025lotus} introduces a simple yet effective adaptation protocol for dense prediction tasks, enabling zero-shot depth and normal estimation. RGB2X~\cite{zeng2024rgbx} and Diffusion Renderer~\cite{liang2025diffusion} are two recent material decomposition and relighting methods, trained on synthetic data, that are direct baselines to us. We additionally include Dsine~\cite{bae2024dsine}, an off-the-shelf normal estimator also trained exclusively on synthetic data.




\paragraph{Synthetic Dataset} 
\label{sec:synthetic dataset}
For evaluation, we curated a subset of 50 objects from the Objaverse Dataset~\cite{objaverse} and rendered their material components—base color, surface normals, and specular maps—using the Blender EEVEE rendering engine\cite{blender}. For each object, we generated 60 turntable views uniformly spaced across 360 degrees. To simulate OLAT-style lighting, we additionally rendered all 346 individual lighting directions per view. Each lighting direction was represented by a custom HDRI environment map constructed using a single hexagonal area light, spatially positioned and oriented to match the corresponding LED in the original Light Stage configuration.

\subsection{Evaluation}

\paragraph{Decomposition} 




We evaluate the effectiveness of our material dataset for inverse decomposition, the task of decomposing an input RGB image into its intrinsic physical components: diffuse albedo, surface normal, and specular albedo. RGB2X and Diffusion Renderer Inverse (DR-IR) are benchmarked on our dataset and the synthetic Objaverse dataset (Sec.~\ref{sec:synthetic dataset}). Lotus-D and Dsine are included for comparison on surface-normal prediction only. Two input lighting setups are considered: \textbf{1) HDRI lighting}, rendered directly for synthetic data and synthesized for our Light Stage dataset by linearly combining OLAT captures using weights derived from the lighting augmentation module; and \textbf{2) OLAT lighting}, where synthetic data are rendered with the prepared 346 OLAT HDRIs (Sec.~\ref{sec:synthetic dataset}) and real data use the captured OLAT images directly. RGB2X is further post-trained on our dataset to evaluate real-world adaptation.

Figure~\ref{fig:inverse-objaverse} presents qualitative decompositions under different lighting setups. Table~\ref{tab:inverse_evaluation_gbuffer} summarizes quantitative metrics across 50 objects from our dataset and 50 objects from Objaverse based on four metrics for visual similarity: MSE, PSNR, SSIM and LPIPS \cite{zhang2018unreasonable}. Under \textbf{HDRI lighting}, our model markedly improves decomposition accuracy. These gains highlight the benefit of real-world polarization cues for disentangling diffuse and geometric components under complex illumination. Under \textbf{OLAT lighting}, where inputs are illuminated by calibrated directional HDRIs, our model also achieves the lowest overall MSE and highest PSNR, confirming robustness to directional illumination. The improvement on specular is less evident, partly because specular reflectance is defined inconsistently across rendering engines—some model it via index of refraction or roughness, whereas our dataset measures it through polarization-based separation, resulting in a different normalization. Despite the smaller numerical gain, qualitative results show clearer highlight separation.



\paragraph{Relighting}




We further evaluate the effectiveness of our polarized reflectance dataset for relighting, where the goal is to render from decomposed material components under novel illumination settings. Two workflows are examined: \textbf{1) the PBR workflow}, which uses diffuse albedo and surface normal for physically based shading, and \textbf{2) the polarization workflow}, which leverages polarization-separated components to model diffuse–specular interactions. To guide the relighting, we additionally provide an irradiance map computed as the integrated $\sum (n \cdot \mathbf{\omega})$ over all sampled light directions weighted by it's color.

We benchmark RGB2X and Diffusion Renderer on both our Light Stage and Objaverse datasets under HDRI and OLAT lighting. RGB2X is further post-trained on our dataset to assess adaptation to real-world reflectance.

Figure~\ref{fig:relighting-lightstage} and Figure~\ref{fig:relighting-objaverse} visualize the relighting results on both datasets. Quantitative comparisons are reported in in Table~\ref{tab:forward_evaluation}, following the same evaluation scale as the decomposition evaluaton. For each lighting setup, we report both PBR and polarization workflows to analyze the influence of physical supervision.

In the \textbf{PBR workflow}, our relighting model produces markedly improved shading continuity and global color balance. On real Light Stage objects, shadows align more faithfully with geometry, and highlight energy remains stable across lighting directions. Quantitatively, PSNR increases from 18.5 dB to 27.8 dB and SSIM from 0.51 to 0.90, confirming that real polarization-aligned data help the network learn a smoother irradiance response. Compared to DR-FR, which tends to overfit synthetic radiance distributions, the post-trained model generalizes better to real illumination, maintaining global color balance and shading smoothness.

In the \textbf{polarization workflow}, relighting results demonstrate stronger lighting consistency and reduced noise in glossy regions. The polarization-based supervision constrains diffuse–specular interaction and mitigates cross-polarization artifacts, yielding more stable illumination transitions on real captures. However, the visual quality on synthetic examples is slightly lower than the PBR workflow—an expected outcome given the domain gap between polarization-separated captures and renderer-generated materials. Despite this, the polarization workflow maintains high physical consistency and perceptual alignment, with LPIPS = 0.200 indicating faithful reconstruction of lighting effects.

\paragraph{Sparse 3D Reconstruction} 

Sparse-view 3D reconstruction is highly sensitive to view-dependent specularities that violate the Lambertian assumption, often leading to inconsistent geometry and photometric errors. Polarization helps address this issue by physically separating diffuse and specular components, enabling cleaner multiview correspondences. However, in real-world scenarios, polarized images are rarely available. With our dataset and proposed polarization workflow, once an inverse model is trained with polarized supervision, it can predict polarization-equivalent outputs, such as diffuse-like images, from standard unpolarized inputs. 

Figure~\ref{fig:reconstruction-lightstage} visualizes how the trained inverse model can help with sparse view reconstruction methods. We showcase the improvement using Dust3r~\cite{dust3r_cvpr24} and Mast3r~\cite{dust3r_cvpr24}, and present quantitative comparisons in Table~\ref{tab:sparse-view-reconstruction}. 

We evaluate three settings, all based on the same 8-view captures on 50 randomly selected objects: (1) using the raw images as input, (2) using diffuse-only images generated by our pretrained inverse model through the polarization workflow, and (3) using diffuse albedo generated by our petrained invese model through the PBR workflow as input. The differences among these experimental settings are illustrated in the top diagram of Figure~\ref{fig:reconstruction-lightstage}. Both Dust3r and Mast3r reconstruct 3D geometry and reproject the recovered points into image space to compute reconstruction errors, summarized in Table~\ref{tab:sparse-view-reconstruction}.

Across these configurations, we also report masked results following~\cite{kirillov2023segment}. The predicted albedo images from our inverse model (setting 3) achieve the highest image-based reconstruction metrics, followed by the predicted diffuse images (setting 2).  When masking is applied, the reconstruction performance with predicted diffuse images improves further. These findings highlight that, once trained, the inverse model effectively generates normalized, specular-free multiview images suitable for downstream 3D reconstruction.  Moreover, the inherent denoising process of the diffusion model contributes to smoother and more consistent reconstructions, the predicted diffuse images even achieve higher PSNR values than the captured diffuse images.

\section{Conclusion and Limitation}

We introduced the large-scale polarized reflection and material dataset of real-world objects at scale, providing dense diffuse–specular separation and material attributes across 200+ objects, 346 lighting directions, and 8 calibrated views. Experiments show that models trained on our measurements achieve substantially improved intrinsic decomposition, relighting quality, and multiview consistency compared to synthetic-only baselines. \textbf{Limitations}. Our capture setup is constrained to static objects within a controlled Light Stage, making highly transparent, dynamic, or strongly anisotropic materials difficult to measure. The dataset does not capture subsurface parameters. Future work will extend the capture to more complex materials and explore scalable, in-the-wild acquisition.

\newpage
{
    \small
    \bibliographystyle{ieeenat_fullname}
    \bibliography{main}

@String(CVPR= {IEEE Conf. Comput. Vis. Pattern Recog.})

@String(ECCV= {Eur. Conf. Comput. Vis.})

@String(TOG= {ACM Trans. Graph.})

@String(ICLR = {Int. Conf. Learn. Represent.})

@String(CVPR  = {CVPR})

@String(ECCV  = {ECCV})

@String(TOG   = {ACM TOG})

@String(ICLR  = {ICLR})

@inproceedings{liang2025diffusion,
  title={Diffusion Renderer: Neural Inverse and Forward Rendering with Video Diffusion Models},
  author={Liang, Ruofan and Gojcic, Zan and Ling, Huan and Munkberg, Jacob and Hasselgren, Jon and Lin, Chih-Hao and Gao, Jun and Keller, Alexander and Vijaykumar, Nandita and Fidler, Sanja and others},
  booktitle={Proceedings of the Computer Vision and Pattern Recognition Conference},
  pages={26069--26080},
  year={2025}
}

@inproceedings{zeng2024rgbx,
  title     = {RGBX: Image Decomposition and Synthesis Using Material- and Lighting-Aware Diffusion Models},
  author    = {Zeng, Zhixin and Deschaintre, Valentin and Georgiev, Iliyan and Hold-Geoffroy, Yannick and Hu, Yuanming and Luan, Fujun and Yan, Ling-Qi and Ha{\v{s}}an, Milo{\v{s}}},
  booktitle = {ACM SIGGRAPH 2024 Conference Proceedings},
  year      = {2024},
  publisher = {Association for Computing Machinery},
  address   = {New York, NY, USA},
  articleno = {75},
  doi       = {10.1145/3641519.3657445},
  url       = {https://research.adobe.com/publication/rgb%E2%86%94x-image-decomposition-and-synthesis-using-material-and-lighting-aware-diffusion-models/}
}

@article{bhattad2023stylegan,
  title     = {StyleGAN Knows Normal, Depth, Albedo, and More},
  author    = {Bhattad, Anand and McKee, Daniel and Hoiem, Derek and Forsyth, David A.},
  journal   = {arXiv preprint arXiv:2301.04638},
  year      = {2023},
  url       = {https://arxiv.org/abs/2301.04638}
}

@inproceedings{li2021openrooms,
  title={Openrooms: An open framework for photorealistic indoor scene datasets},
  author={Li, Zhengqin and Yu, Ting-Wei and Sang, Shen and Wang, Sarah and Song, Meng and Liu, Yuhan and Yeh, Yu-Ying and Zhu, Rui and Gundavarapu, Nitesh and Shi, Jia and others},
  booktitle={Proceedings of the IEEE/CVF conference on computer vision and pattern recognition},
  pages={7190--7199},
  year={2021}
}

@inproceedings{wimbauer2022rendering,
  title={De-rendering 3d objects in the wild},
  author={Wimbauer, Felix and Wu, Shangzhe and Rupprecht, Christian},
  booktitle={Proceedings of the IEEE/CVF Conference on Computer Vision and Pattern Recognition},
  pages={18490--18499},
  year={2022}
}

@inproceedings{
he2025lotus,
title={Lotus: Diffusion-based Visual Foundation Model for High-quality Dense Prediction},
author={Jing He and Haodong Li and Wei Yin and Yixun Liang and Leheng Li and Kaiqiang Zhou and Hongbo Zhang and Bingbing Liu and Ying-Cong Chen},
booktitle={The Thirteenth International Conference on Learning Representations},
year={2025},
url={https://openreview.net/forum?id=stK7iOPH9Q}
}

@misc{ke2025marigold,
  title={Marigold: Affordable Adaptation of Diffusion-Based Image Generators for Image Analysis},
  author={Bingxin Ke and Kevin Qu and Tianfu Wang and Nando Metzger and Shengyu Huang and Bo Li and Anton Obukhov and Konrad Schindler},
  year={2025},
  eprint={2505.09358},
  archivePrefix={arXiv},
  primaryClass={cs.CV}
}

@article{objaverse,
  title={Objaverse: A Universe of Annotated 3D Objects},
  author={Matt Deitke and Dustin Schwenk and Jordi Salvador and Luca Weihs and
          Oscar Michel and Eli VanderBilt and Ludwig Schmidt and
          Kiana Ehsani and Aniruddha Kembhavi and Ali Farhadi},
  journal={arXiv preprint arXiv:2212.08051},
  year={2022}
}

@article{objaverseXL,
  title={Objaverse-XL: A Universe of 10M+ 3D Objects},
  author={Matt Deitke and Ruoshi Liu and Matthew Wallingford and Huong Ngo and
          Oscar Michel and Aditya Kusupati and Alan Fan and Christian Laforte and
          Vikram Voleti and Samir Yitzhak Gadre and Eli VanderBilt and
          Aniruddha Kembhavi and Carl Vondrick and Georgia Gkioxari and
          Kiana Ehsani and Ludwig Schmidt and Ali Farhadi},
  journal={arXiv preprint arXiv:2307.05663},
  year={2023}
}

@inproceedings{murmann2019dataset,
  title={A dataset of multi-illumination images in the wild},
  author={Murmann, Lukas and Gharbi, Michael and Aittala, Miika and Durand, Fredo},
  booktitle={Proceedings of the IEEE/CVF International Conference on Computer Vision},
  pages={4080--4089},
  year={2019}
}

@inproceedings{zhu2022learning,
    author = {Zhu, Jingsen and Luan, Fujun and Huo, Yuchi and Lin, Zihao and Zhong, Zhihua and Xi, Dianbing and Wang, Rui and Bao, Hujun and Zheng, Jiaxiang and Tang, Rui},
    title = {Learning-Based Inverse Rendering of Complex Indoor Scenes with Differentiable Monte Carlo Raytracing},
    year = {2022},
    publisher = {ACM},
    url = {https://doi.org/10.1145/3550469.3555407},
    booktitle = {SIGGRAPH Asia 2022 Conference Papers},
    articleno = {6},
    numpages = {8}
}

@inproceedings{roberts2021hypersim,
  title={Hypersim: A photorealistic synthetic dataset for holistic indoor scene understanding},
  author={Roberts, Mike and Ramapuram, Jason and Ranjan, Anurag and Kumar, Atulit and Bautista, Miguel Angel and Paczan, Nathan and Webb, Russ and Susskind, Joshua M},
  booktitle={Proceedings of the IEEE/CVF international conference on computer vision},
  pages={10912--10922},
  year={2021}
}

@inproceedings{dust3r_cvpr24,
      title={DUSt3R: Geometric 3D Vision Made Easy}, 
      author={Shuzhe Wang and Vincent Leroy and Yohann Cabon and Boris Chidlovskii and Jerome Revaud},
      booktitle = {CVPR},
      year = {2024}
}

@inproceedings{Luo:2024:Correlation-aware,
  title={Correlation-aware Encoder-Decoder with Adapters for SVBRDF Acquisition},
  author={Di Luo and Hanxiao Sun and Lei Ma and Jian Yang and Beibei Wang},
  booktitle={Proceedings of SIGGRAPH Asia 2024},
  year={2024},
}

@article{barrow1978recovering,
  title={Recovering intrinsic scene characteristics},
  author={Barrow, Harry and Tenenbaum, J and Hanson, A and Riseman, E},
  journal={Comput. vis. syst},
  volume={2},
  number={3-26},
  pages={2},
  year={1978}
}

@inproceedings{chen2013simple,
  title={A simple model for intrinsic image decomposition with depth cues},
  author={Chen, Qifeng and Koltun, Vladlen},
  booktitle={Proceedings of the IEEE international conference on computer vision},
  pages={241--248},
  year={2013}
}

@inproceedings{garces2012intrinsic,
  title={Intrinsic images by clustering},
  author={Garces, Elena and Munoz, Adolfo and Lopez-Moreno, Jorge and Gutierrez, Diego},
  booktitle={Computer graphics forum},
  volume={31},
  number={4},
  pages={1415--1424},
  year={2012},
  organization={Wiley Online Library}
}

@article{zhao2012closed,
  title={A closed-form solution to retinex with nonlocal texture constraints},
  author={Zhao, Qi and Tan, Ping and Dai, Qiang and Shen, Li and Wu, Enhua and Lin, Stephen},
  journal={IEEE transactions on pattern analysis and machine intelligence},
  volume={34},
  number={7},
  pages={1437--1444},
  year={2012},
  publisher={IEEE}
}

@article{bell2014intrinsic,
  title={Intrinsic images in the wild},
  author={Bell, Sean and Bala, Kavita and Snavely, Noah},
  journal={ACM Transactions on Graphics (TOG)},
  volume={33},
  number={4},
  pages={1--12},
  year={2014},
  publisher={ACM New York, NY, USA}
}

@article{rother2011recovering,
  title={Recovering intrinsic images with a global sparsity prior on reflectance},
  author={Rother, Carsten and Kiefel, Martin and Zhang, Lumin and Sch{\"o}lkopf, Bernhard and Gehler, Peter},
  journal={Advances in neural information processing systems},
  volume={24},
  year={2011}
}

@article{meka2021real,
  title={Real-time global illumination decomposition of videos},
  author={Meka, Abhimitra and Shafiei, Mohammad and Zollh{\"o}fer, Michael and Richardt, Christian and Theobalt, Christian},
  journal={ACM Transactions on Graphics (ToG)},
  volume={40},
  number={3},
  pages={1--16},
  year={2021},
  publisher={ACM New York, NY}
}

@inproceedings{choi2023ibl,
  title={IBL-NeRF: Image-Based Lighting Formulation of Neural Radiance Fields},
  author={Choi, Changwoon and Kim, Juhyeon and Kim, Young Min},
  booktitle={Computer Graphics Forum},
  volume={42},
  number={7},
  pages={e14929},
  year={2023},
  organization={Wiley Online Library}
}

@article{careaga2023intrinsic,
  title={Intrinsic image decomposition via ordinal shading},
  author={Careaga, Chris and Aksoy, Ya{\u{g}}{\i}z},
  journal={ACM Transactions on Graphics},
  volume={43},
  number={1},
  pages={1--24},
  year={2023},
  publisher={ACM New York, NY, USA}
}

@inproceedings{zhou2015learning,
  title={Learning data-driven reflectance priors for intrinsic image decomposition},
  author={Zhou, Tinghui and Krahenbuhl, Philipp and Efros, Alexei A},
  booktitle={Proceedings of the IEEE international conference on computer vision},
  pages={3469--3477},
  year={2015}
}

@inproceedings{fan2018revisiting,
  title={Revisiting deep intrinsic image decompositions},
  author={Fan, Qingnan and Yang, Jiaolong and Hua, Gang and Chen, Baoquan and Wipf, David},
  booktitle={Proceedings of the IEEE conference on computer vision and pattern recognition},
  pages={8944--8952},
  year={2018}
}

@inproceedings{wang2021learning,
  title={Learning indoor inverse rendering with 3d spatially-varying lighting},
  author={Wang, Zian and Philion, Jonah and Fidler, Sanja and Kautz, Jan},
  booktitle={Proceedings of the IEEE/CVF International Conference on Computer Vision},
  pages={12538--12547},
  year={2021}
}

@inproceedings{shi2017learning,
  title={Learning non-lambertian object intrinsics across shapenet categories},
  author={Shi, Jian and Dong, Yue and Su, Hao and Yu, Stella X},
  booktitle={Proceedings of the IEEE conference on computer vision and pattern recognition},
  pages={1685--1694},
  year={2017}
}

@inproceedings{narihira2015direct,
  title={Direct intrinsics: Learning albedo-shading decomposition by convolutional regression},
  author={Narihira, Takuya and Maire, Michael and Yu, Stella X},
  booktitle={Proceedings of the IEEE international conference on computer vision},
  pages={2992--2992},
  year={2015}
}

@inproceedings{boss2020two,
  title={Two-shot spatially-varying brdf and shape estimation},
  author={Boss, Mark and Jampani, Varun and Kim, Kihwan and Lensch, Hendrik and Kautz, Jan},
  booktitle={Proceedings of the IEEE/CVF Conference on Computer Vision and Pattern Recognition},
  pages={3982--3991},
  year={2020}
}

@inproceedings{luo2024intrinsicdiffusion,
  title={Intrinsicdiffusion: Joint intrinsic layers from latent diffusion models},
  author={Luo, Jundan and Ceylan, Duygu and Yoon, Jae Shin and Zhao, Nanxuan and Philip, Julien and Fr{\"u}hst{\"u}ck, Anna and Li, Wenbin and Richardt, Christian and Wang, Tuanfeng},
  booktitle={ACM SIGGRAPH 2024 Conference Papers},
  pages={1--11},
  year={2024}
}

@inproceedings{chen2025uni,
  title={Uni-Renderer: Unifying Rendering and Inverse Rendering Via Dual Stream Diffusion},
  author={Chen, Zhifei and Xu, Tianshuo and Ge, Wenhang and Wu, Leyi and Yan, Dongyu and He, Jing and Wang, Luozhou and Zeng, Lu and Zhang, Shunsi and Chen, Ying-Cong},
  booktitle={Proceedings of the Computer Vision and Pattern Recognition Conference},
  pages={26504--26513},
  year={2025}
}

@inproceedings{choi2025channel,
  title={Channel-wise Noise Scheduled Diffusion for Inverse Rendering in Indoor Scenes},
  author={Choi, JunYong and Sagong, Min-cheol and Lee, SeokYeong and Jung, Seung-Won and Kim, Ig-Jae and Cho, Junghyun},
  booktitle={Proceedings of the Computer Vision and Pattern Recognition Conference},
  pages={5773--5782},
  year={2025}
}

@inproceedings{zheng2025dnf,
  title={DNF-Intrinsic: Deterministic Noise-Free Diffusion for Indoor Inverse Rendering},
  author={Zheng, Rongjia and Zhang, Qing and Long, Chengjiang and Zheng, Wei-Shi},
  booktitle={Proceedings of the IEEE/CVF International Conference on Computer Vision},
  pages={10342--10352},
  year={2025}
}

@inproceedings{sun2025ouroboros,
  title={Ouroboros: Single-step Diffusion Models for Cycle-consistent Forward and Inverse Rendering},
  author={Sun, Shanlin and Wang, Yifan and Zhang, Hanwen and Xiong, Yifeng and Ren, Qin and Fang, Ruogu and Xie, Xiaohui and You, Chenyu},
  booktitle={Proceedings of the IEEE/CVF International Conference on Computer Vision},
  pages={10386--10397},
  year={2025}
}

@inproceedings{li2018cgintrinsics,
  title     = {CGIntrinsics: Better Intrinsic Image Decomposition through Physically-Based Rendering},
  author    = {Li, Zhengqi and Snavely, Noah},
  booktitle = {Proceedings of the European Conference on Computer Vision (ECCV)},
  year      = {2018},
  pages     = {371--387},
  doi       = {10.1007/978-3-030-01264-9_23}
}

@article{li2018interiornet,
  title={Interiornet: Mega-scale multi-sensor photo-realistic indoor scenes dataset},
  author={Li, Wenbin and Saeedi, Sajad and McCormac, John and Clark, Ronald and Tzoumanikas, Dimos and Ye, Qing and Huang, Yuzhong and Tang, Rui and Leutenegger, Stefan},
  journal={arXiv preprint arXiv:1809.00716},
  year={2018}
}

@inproceedings{li2020inverse,
  title={Inverse rendering for complex indoor scenes: Shape, spatially-varying lighting and svbrdf from a single image},
  author={Li, Zhengqin and Shafiei, Mohammad and Ramamoorthi, Ravi and Sunkavalli, Kalyan and Chandraker, Manmohan},
  booktitle={Proceedings of the IEEE/CVF conference on computer vision and pattern recognition},
  pages={2475--2484},
  year={2020}
}

@inproceedings{ramamoorthi2001signal,
  title={A signal-processing framework for inverse rendering},
  author={Ramamoorthi, Ravi and Hanrahan, Pat},
  booktitle={Proceedings of the 28th annual conference on Computer graphics and interactive techniques},
  pages={117--128},
  year={2001}
}

@inproceedings{kirillov2023segment,
  title={Segment anything},
  author={Kirillov, Alexander and Mintun, Eric and Ravi, Nikhila and Mao, Hanzi and Rolland, Chloe and Gustafson, Laura and Xiao, Tete and Whitehead, Spencer and Berg, Alexander C and Lo, Wan-Yen and others},
  booktitle={Proceedings of the IEEE/CVF international conference on computer vision},
  pages={4015--4026},
  year={2023}
}

@article{hu2022lora,
  title={Lora: Low-rank adaptation of large language models.},
  author={Hu, Edward J and Shen, Yelong and Wallis, Phillip and Allen-Zhu, Zeyuan and Li, Yuanzhi and Wang, Shean and Wang, Lu and Chen, Weizhu and others},
  journal={ICLR},
  volume={1},
  number={2},
  pages={3},
  year={2022}
}

@misc{blender,
  title        = {Blender - a 3D modelling and rendering package},
  author       = {{Blender Online Community}},
  publisher    = {Blender Foundation},
  howpublished = {\url{https://www.blender.org}},
  year         = {2024}
}

@article{10.1145/3618358,
author = {Ma, Xiaohe and Xu, Xianmin and Zhang, Leyao and Zhou, Kun and Wu, Hongzhi},
title = {OpenSVBRDF: A Database of Measured Spatially-Varying Reflectance},
year = {2023},
issue_date = {December 2023},
publisher = {Association for Computing Machinery},
address = {New York, NY, USA},
volume = {42},
number = {6},
issn = {0730-0301},
url = {https://doi.org/10.1145/3618358},
doi = {10.1145/3618358},
journal = {ACM Trans. Graph.},
month = dec,
articleno = {254},
numpages = {14}
}

@inproceedings{zhang2018unreasonable,
  title={The unreasonable effectiveness of deep features as a perceptual metric},
  author={Zhang, Richard and Isola, Phillip and Efros, Alexei A and Shechtman, Eli and Wang, Oliver},
  booktitle={Proceedings of the IEEE conference on computer vision and pattern recognition},
  pages={586--595},
  year={2018}
}

@inproceedings{bae2024dsine,
    title     = {Rethinking Inductive Biases for Surface Normal Estimation},
    author    = {Gwangbin Bae and Andrew J. Davison},
    booktitle = {IEEE/CVF Conference on Computer Vision and Pattern Recognition (CVPR)},
    year      = {2024}
}

@misc{liu2024openillumination,
    title={OpenIllumination: A Multi-Illumination Dataset for Inverse Rendering Evaluation on Real Objects}, 
    author={Isabella Liu and Linghao Chen and Ziyang Fu and Liwen Wu and Haian Jin and Zhong Li and Chin Ming Ryan Wong and Yi Xu and Ravi Ramamoorthi and Zexiang Xu and Hao Su},
    year={2024},
    eprint={2309.07921},
    archivePrefix={arXiv},
    primaryClass={cs.CV}
}
}


\end{document}